\documentclass[10pt]{article}
\usepackage[utf8]{inputenc}
\usepackage{textcomp}

\usepackage[a4paper,margin=2cm]{geometry}
\usepackage{graphicx} 
\usepackage{tabularx} 
\usepackage{array}
\usepackage{booktabs}
\usepackage{graphicx}
\usepackage{tcolorbox}
\usepackage{makecell}
\usepackage{enumitem}
\usepackage{amsmath} 
\usepackage{changepage}
\usepackage{caption}
\usepackage{subcaption}
\usepackage[colorlinks=true,linkcolor=blue,citecolor=blue,urlcolor=blue]{hyperref}

\title{BMGQ: A Bottom-up Method for Generating Complex Multi-hop Reasoning Questions from Semi-structured Data}
\author{\footnotesize
Bingsen Qiu, Zijian Liu, Xiao Liu, Bingjie Wang, Feier Zhang, Yixuan Qin, Chunyan Li, Haoshen Yang, Zeren Gao\\[0.5ex] 
\small
ByteDance DMC\thanks{ByteDance Data and Model Service Center, Data Exploration Team} \\[0.5ex]
\footnotesize
\texttt{\{qiubingsen.123, liuzijian.109, liuxiao.0423, wangbingjie, zhangfeier, qinyixuan.1, lichunyan.ivy\}@bytedance.com}\\[0.5ex]
\footnotesize
\texttt{\{yanghaoshen, gaozeren\}@jiyunhudong.com}
}
\date{\small \today}

\begin{document}

\maketitle
\begin{abstract}
Building training-ready multi-hop question answering (QA) datasets that truly stress a model’s retrieval and reasoning abilities remains highly challenging recently. While there have been a few recent evaluation datasets that capture the characteristics of hard-to-search but easy-to-verify problems—requiring the integration of ambiguous, indirect, and cross-domain cues—these data resources remain scarce and are mostly designed for evaluation, making them unsuitable for supervised fine-tuning (SFT) or reinforcement learning (RL). Meanwhile, manually curating non-trivially retrievable questions—where answers cannot be found through a single direct query but instead require multi-hop reasoning over oblique and loosely connected evidence—incurs prohibitive human costs and fails to scale, creating a critical data bottleneck for training high-capability retrieval-and-reasoning agents.

To address this, we present \textbf{BMGQ}, a bottom-up automated method for generating high-difficulty, training-ready multi-hop questions from semi-structured knowledge sources. The BMGQ system (i) grows diverse, logically labeled evidence clusters through Natural Language Inference (NLI)-based relation typing and diversity-aware expansion; (ii) applies reverse question construction to compose oblique cues so that isolated signals are underinformative but their combination uniquely identifies the target entity; and (iii) enforces quality with a two-step evaluation pipeline that combines multi-model consensus filtering with structured constraint decomposition and evidence-based matching. The result is a scalable process that yields complex, retrieval-resistant yet verifiable questions suitable for SFT/RL training as well as challenging evaluation—substantially reducing human curation effort while preserving the difficulty profile of strong evaluation benchmarks.
\end{abstract}

\section{Introduction}
Large language models (LLMs) have achieved remarkable progress across many natural language processing tasks, yet their ability to conduct deep research involving multi-hop retrieval and reasoning remains limited. This capability is crucial for solving problems where answers cannot be found through a single lookup but require integrating multiple pieces of indirect, cross-domain evidence. While existing multi-hop QA training datasets provide useful testbeds, most of them\cite{yang2018hotpotqa}\cite{talmor2018repartitioning} rely on relatively shallow reasoning chains, offering limited value for training models that can handle genuinely complex retrieval and reasoning tasks.

In response to this gap, a new line of evaluation benchmarks has emerged, specifically targeting challenging deep reasoning scenarios. These datasets are intentionally designed to be hard to search and easy to verify: individual clues are vague and insufficient to reach the answer, but when combined, they collapse the search space to a single, verifiable entity. While such benchmarks effectively capture the difficulty profile required to improve LLM reasoning capabilities, they are expensive to construct and cannot be used for large-scale SFT or RL.

To overcome these scalability limitations, recent work has explored automatic question generation pipelines. These approaches demonstrate the feasibility of synthesizing multi-hop questions at scale, but most remain limited in depth: they often involve only a few reasoning steps, lack complex cross-domain structures, and do not enforce strict verification of answer uniqueness. As a result, the generated questions are still far from the difficulty and reliability required for training powerful reasoning agents.

Motivated by these gaps, we introduce \textbf{BMGQ}, a bottom-up automated framework for constructing high-difficulty multi-hop datasets suitable for both training and evaluation. BMGQ transforms semi-structured knowledge sources into structured evidence clusters, builds diverse and logically coherent evidence clusters, and employs a reverse construction strategy that composes indirect clues to ensure that the answer is uniquely identifiable only when all clues are considered together. Finally, a dedicated quality evaluation system combines multi-model consensus filtering with structured constraint decomposition and evidence-based matching, ensuring that each retained question is both challenging and verifiably correct.

By combining the difficulty profile of high-end evaluation benchmarks with the scalability of automated synthesis, BMGQ enables the creation of large, training-ready multi-hop datasets that are capable of truly stressing and improving the deep retrieval and reasoning abilities of large language models. To support reproducibility, we release a partial subset of the BMGQ dataset, including only the final questions and their seed answers, at
https://huggingface.co/datasets/Fayer/BMGQ-MultiHop-Sample

\section{Related Works}
\subsection{Constructing Deep, Ambiguous, and Uniquely Solvable Multi-Hop Questions}
\label{sec:problem_definition}
We study the problem of automatically constructing large-scale multi-hop QA datasets that match the reasoning difficulty of human-curated benchmarks such as BrowseComp while ensuring strict answer uniqueness. 
Unlike traditional QA tasks, the goal is not merely to answer a question, but to generate a question that:
\begin{enumerate}
    \item Requires deep multi-step reasoning across a chain of heterogeneous evidence.
    \item Contains ambiguous or indirect clues that prevent shallow retrieval or direct lookup.
    \item Remains uniquely solvable, meaning only one entity in the evidence space satisfies all constraints.
    \item Is verifiable against structured evidence, allowing downstream models to check correctness.
\end{enumerate}

We formalize the dataset construction problem as generating a natural-language question $Q$ from a structured evidence graph $G=(V,E)$ induced by a seed entity $s$ -- which also serves as the answer $A$ to be recovered. Each question imposes a set of constraints $C(Q)$ over nodes and attributes in $G$. The target answer $A$ must satisfy all constraints:
\[
A = \{\, v \in V \mid v \models C(Q) \, \}.
\]

To ensure strict answer uniqueness, the question must satisfy:
\[
|A| = 1.
\]

At the same time, the reasoning path required to reach $A$ must be non-trivial:
\[
\text{depth}(C(Q)) \ge n,
\]
with $n$ representing a configurable minimal clue depth.

This formulation highlights why automatic construction is difficult: deep multi-hop reasoning, clue ambiguity, cross-domain transitions, and strict uniqueness must be satisfied \emph{simultaneously}, yet existing datasets or generation frameworks do not jointly enforce these constraints.

\subsection{Existing Benchmarks for Multi-Hop Reasoning}
Building directly upon the problem defined in Section~\ref{sec:problem_definition}, we evaluate how existing multi-hop QA benchmarks relate to the goal of constructing deep, ambiguous, and uniquely solvable questions. Although these datasets have significantly advanced multi-hop reasoning research, none of them are designed to satisfy all required properties—deep reasoning depth, indirect clue construction, retrieval difficulty, and strict answer uniqueness—simultaneously.

Early benchmarks such as HotpotQA\cite{yang2018hotpotqa}, ComplexWebQuestions\cite{talmor2018repartitioning}, MuSiQue\cite{trivedi2022musique}, QASC\cite{khot2020qasc}, and 2WikiMultiHopQA \cite{ho2020constructing} laid the groundwork for multi-hop QA. HotpotQA introduced multi-document reasoning with annotated supporting sentences, but its reasoning chains are typically shallow and often solvable via lexical matching rather than true multi-step inference. ComplexWebQuestions expanded compositional complexity through knowledge-graph path generation, yet its template-based design limits linguistic diversity and yields predictable reasoning structures. MuSiQue aimed to mitigate shortcut exploitation by composing single-hop facts into longer chains, but the majority of its questions remain short and lack enforced retrieval difficulty. Domain-focused and hybrid-structured datasets such as QASC and 2WikiMultiHopQA broadened coverage but still depend heavily on explicit facts or templates. Collectively, these benchmarks remain limited by shallow reasoning, explicit clues, and a lack of guarantee on answer uniqueness.

A more recent and significantly harder benchmark, BrowseComp\cite{wei2025browsecomp}, represents a distinct shift in design philosophy. BrowseComp constructs questions that are deliberately vague when examined clue-by-clue, forcing deep multi-hop retrieval across multiple domains, yet the final answer is short and objectively verifiable. This “hard-to-search, easy-to-verify” paradigm better captures the challenges of real-world deep research tasks, requiring models to integrate scattered, indirect signals rather than relying on direct lookup. However, despite its conceptual advances, our internal manual evaluation suggests that many BrowseComp questions still exhibit imperfections in uniqueness: multiple plausible answers may satisfy the same fuzzy constraints. Moreover, BrowseComp is intentionally released as an evaluation benchmark, not a scalable training corpus, making it unsuitable for supervised finetuning or reinforcement learning.

In summary, while existing multi-hop datasets collectively push the field toward more realistic reasoning, they fall short of the requirements outlined in Section~\ref{sec:problem_definition}. Early benchmarks lack depth, ambiguity, and retrieval difficulty, whereas BrowseComp achieves substantial difficulty but still does not guarantee strict answer uniqueness or provide a method for scalable dataset construction. This persistent gap highlights the need for a principled construction framework capable of producing deep-reasoning, ambiguous-yet-uniquely-solvable questions at scale—an objective that motivates the methodology presented in this work.

\subsection{Automatic Multi-Hop Dataset Construction Frameworks}
Given the limitations of existing multi-hop benchmarks, researchers have increasingly explored automatic question construction as a scalable alternative to manual curation. Early efforts primarily relied on rule-based templates or knowledge graph traversal, aiming to generate multi-hop questions by expanding relations in structured datasets. For example, ComplexWebQuestions constructs compositional questions by extending WebQuestionsSP using Freebase relation paths. Although this approach enables large-scale synthesis, its dependence on fixed templates results in predictable linguistic forms and shallow, deterministic reasoning chains. Similar KG-driven pipelines—such as those used in 2WikiMultiHopQA -- combine Wikidata triples with supporting Wikipedia passages, but remain constrained by explicit relation paths that lack the ambiguity and retrieval difficulty required in deep multi-hop contexts.

To move beyond template rigidity, several works have explored retrieval-augmented generation. The most influential of these is HopWeaver \cite{shen2025hopweaver}, which synthesizes bridge and comparison questions by identifying semantically connected entity pairs, retrieving intermediate passages from Wikipedia, and prompting an LLM to compose a question integrating those hops. HopWeaver demonstrates that automatic multi-hop generation can leverage textual evidence and large language models to produce more natural questions than purely symbolic pipelines. However, its reasoning patterns remain limited to a narrow set of hop types (mainly bridge and comparison), and most generated questions only require two to three hops. Moreover, HopWeaver does not explicitly enforce controlled ambiguity or guarantee strict answer uniqueness—properties that are central to the construction of hard-to-search yet uniquely solvable questions. As a result, although HopWeaver marks an important step forward, it still falls short of producing BrowseComp-level difficulty in a scalable and principled way.

Taken together, existing automatic construction methods provide valuable insights but do not jointly address the core challenges required for generating deep multi-hop reasoning datasets. They either rely on shallow templates, produce limited reasoning diversity, or lack robust verification pipelines to ensure correctness and uniqueness. These gaps motivate the development of a new construction framework—one that leverages structured evidence graphs, bottom-up reverse synthesis, and explicit quality evaluation—to generate BrowseComp-level questions at scale.

\subsection{Remaining Gaps and Our Contribution}
From the above analysis, we identify a critical gap:
there exists no automatic pipeline capable of generating BrowseComp-level hard multi-hop questions with explicit, auditable evidence and guaranteed answer uniqueness. To bridge this gap, we propose a new automated construction framework that integrates:
\begin{itemize}
    \item structured evidence graph building,
    \item depth-controlled clue selection,
    \item bottom-up question synthesis,
    \item graph-based structural sanity checks, and
    \item a two-layer Data Quality Evaluation System combining 
    (i) model-based coarse filtering and 
    (ii) predicate-level constraint verification.
\end{itemize}

Formally, the goal of our system is:
\[
\text{Construct}(Q) \quad \text{s.t.} \quad
\begin{cases}
\text{depth}(C(Q)) \ge n, \\
\text{ambiguity}(Q) \ge \tau_1, \\
\text{retrieval\text{-}difficulty}(Q) \ge \tau_2, \\
\bigl| \{\, v \in V \mid v \models C(Q) \,\} \bigr| = 1.
\end{cases}
\]

This framework unifies graph construction, cue abstraction, question synthesis, and evidence-based verification. It enables the construction of BrowseComp-grade multi-hop reasoning datasets at scale, while ensuring strict answer uniqueness, logical coherence, and structured auditability.

\section{BMGQ Framework: Bottom-up Method for Generating Complex Multi-hop Reasoning Questions}
\begin{figure*}[t]
    \centering
    \includegraphics[width=\textwidth]{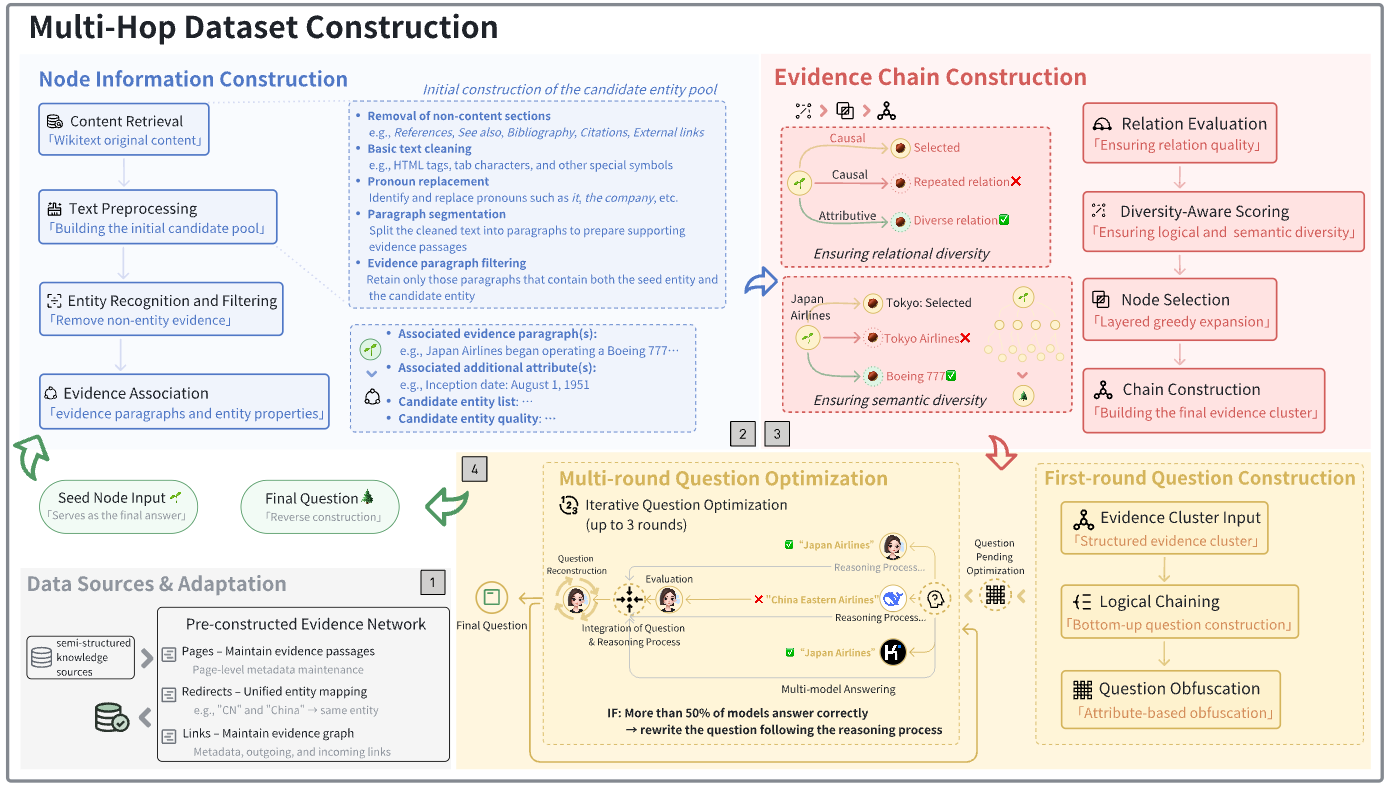}
    \captionsetup{font=small, labelfont=it}
    \caption{ Multi-hop dataset construction framework of our system. The construction process includes: (1) \textbf{Data Sources \& Adaptation}, (2) \textbf{Node Information Construction}, (3) \textbf{Evidence Chain Construction}, and (4) \textbf{Question Construction \& Optimization}. Colored blocks correspond to each stage, with step indices annotated for cross-reference in the main text.}  
    \label{fig:wf_construction}
\end{figure*}

Our methodology for constructing a multi-hop reasoning dataset follows a structured four-stage pipeline, collectively referred to as BMGQ (Bottom-up Method for Generating Complex Multi-hop
Reasoning Questions), as illustrated in Figure~\ref{fig:wf_construction}. In \textbf{Part 1 (Data Sources \& Adaptation)}, semi-structured data are adapted into a lightweight relational database and pre-constructed evidence network. In \textbf{Part 2 (Node Information Construction)}, we retrieve raw page content, perform text preprocessing, remove non-entity evidence, and link candidate entities with their supporting paragraphs. In \textbf{Part 3 (Evidence Chain Construction)}, relation classification guarantees the validity of edges, diversity-aware evaluation mitigates semantic redundancy, and breadth-first expansion yields a balanced and interpretable evidence cluster. Finally, in \textbf{Part 4 (Question Construction \& Optimization)}, a bottom-up reverse generation strategy constructs the initial question, followed by multi-round optimization to enhance abstraction, difficulty, and uniqueness. Together, these components form the core of BMGQ, enabling the automatic generation of high-quality, logically coherent, and challenging multi-hop questions grounded in real-world knowledge.

\subsection{Data Source \& Adaptation}
The construction of our dataset is distributed as large-scale SQL files. While comprehensive, this semi-structured format is not directly suitable for efficient graph queries or multi-hop reasoning. To address this issue, we preprocess the raw data and transform it into a lightweight, high-performance relational database. This design ensures structured organization, millisecond-level query efficiency, and a robust foundation for subsequent reasoning tasks.

\subsection{Node Information Construction}
\label{sec:node_info_construction}
Once the structured database has been established, the next step is to construct an informative contextual environment for a given \textbf{seed entity}\footnote{The target answer of the constructed question.} . This process aims to identify high-quality candidate entities and their corresponding evidence passages, which serve as the foundation for multi-hop evidence chain construction.

The key challenge arises from the fact that a single webpage may contain hundreds or even thousands of outbound links, many of which correspond to generic terms, abstract concepts, or weakly related references. Blindly expanding all such links leads to  \textbf{semantic drift} \cite{yuan2024alleviating}\cite{zhang2019addressing} — a phenomenon where the reasoning path gradually loses topical or contextual relevance, resulting in misleading or meaningless connections. To mitigate this issue, we introduce a multi-stage pipeline that integrates preprocessing, entity recognition, and evidence alignment. An annotated example illustrating the prevalence of irrelevant or weakly connected links in raw Wikipedia pages is provided in Appendix A.

\subsubsection{Goals and Challenges}
\begin{itemize}
    \item \textbf{Objective}: Extract a set of candidate entities (outlinked Wikipedia pages) and their supporting evidence passages from the seed entity’s page.
    \item \textbf{Challenge}: Avoid semantic drift by filtering out irrelevant, overly abstract, or weakly connected terms (e.g., privatised, capital, tons), which otherwise disrupt logical consistency across reasoning chains.
\end{itemize}

For instance, in the entry \textit{Japan Airlines}, some outbound links correspond to irrelevant common nouns (e.g., \textit{mail}, \textit{capital}), whereas others correspond to meaningful entities (e.g., \textit{All Nippon Airways}, \textit{Tokyo}). A full annotated example is provided in Appendix A.

\subsubsection{Processing Pipeline}
\label{sec:processing_pipeline}
The Node Information Construction stage (the part 2 of the Figure~\ref{fig:wf_construction}) builds a structured, semantically coherent neighborhood around each seed entity to support subsequent evidence chain expansion. This process involves lightweight text preprocessing followed by a core filtering step based on named entity recognition (NER).

\begin{enumerate}
    \item \textbf{Text Preprocessing and Candidate Extraction}
    
    Given a seed entity, we retrieve its webpage text content and remove non-essential sections (e.g., “See also”, “References”) and markup artifacts. The cleaned text is segmented into paragraphs, and all internal hyperlinks are extracted as initial candidate entities. These links preserve webpage’s rich contextual connectivity while keeping the preprocessing lightweight.

    \item \textbf{NER-Based Candidate Filtering}
    
    The cornerstone of this stage is robust entity filtering. After evaluating multiple approaches—including rule-based heuristics and statistical sequence models, which are still commonly used as baselines in recent NER research \cite{li2020survey}\cite{yadav2019survey}\cite{wang2020two}—we adopt a transformer-based BERT NER model \cite{devlin2019bert}(dslim/bert-large-NER \cite{devlin2018bert}) due to its superior contextual discrimination.
    
The model reliably separates valid entities from noisy or abstract terms, significantly reducing semantic drift.

\begin{tcolorbox}[colback=gray!5,colframe=black,title=Example 1]
\begin{itemize}
    \item \textbf{Accepted}: \textit{Albert Einstein} (Person), \textit{Pacific Ocean} (Location), \textit{Bytedance Inc} (Organization), \textit{World War II} (Event).
    \item \textbf{Rejected}: \textit{Philosophy} (abstract concept), \textit{Meditation} (general term), \textit{List of countries} (meta-page), \textit{Ocean} (overly broad category).
\end{itemize} 
\end{tcolorbox}

Through a combination of preprocessing, NER-based filtering, and evidence alignment, we construct a high-quality contextual environment around each seed entity. This design minimizes semantic drift and ensures that only semantically grounded, evidence-supported entities progress to the next stage of evidence chain construction.

\item \textbf{Evidence Association and Context Refinement}

For each retained entity, we retrieve its originating paragraph to guarantee textual co-occurrence with the seed entity. We then apply coreference resolution to replace ambiguous mentions with explicit names and optionally enrich entities with structured attributes from Wikidata (e.g., inception date, headquarters, affiliations).

This selective filtering pipeline yields a clean, semantically grounded candidate set and associated evidence passages, minimizing semantic drift and ensuring logical coherence for downstream evidence chain construction.
\end{enumerate}

\subsection{Evidence Chain Construction}
After generating high-quality candidate entities and evidence passages in Section \ref{sec:node_info_construction}, the next stage is to construct evidence chains—multi-hop paths that connect entities through explicit, logically interpretable relationships. The goal is to move beyond surface-level similarity and build reasoning paths that mimic human-style exploration, ensuring both semantic diversity and logical rigor.

The central challenge in constructing evidence chains lies in ensuring \textbf{logical connectivity} rather than mere topical similarity. Pure similarity-based methods risk producing \textbf{semantic homogeneity}, where expansions remain trapped in a single domain. For example, starting from Japan Airlines, semantic similarity expansion tends to generate only other airlines, resulting in horizontally enumerated entities without explanatory depth. Such graphs fail to uncover meaningful relations (e.g., usage of aircraft models, regulatory reforms, or cultural connections) and thus cannot support multi-hop reasoning.

\begin{figure}[htbp]
  \centering
  \begin{subfigure}[t]{0.48\textwidth}
    \centering
    \includegraphics[width=\linewidth]{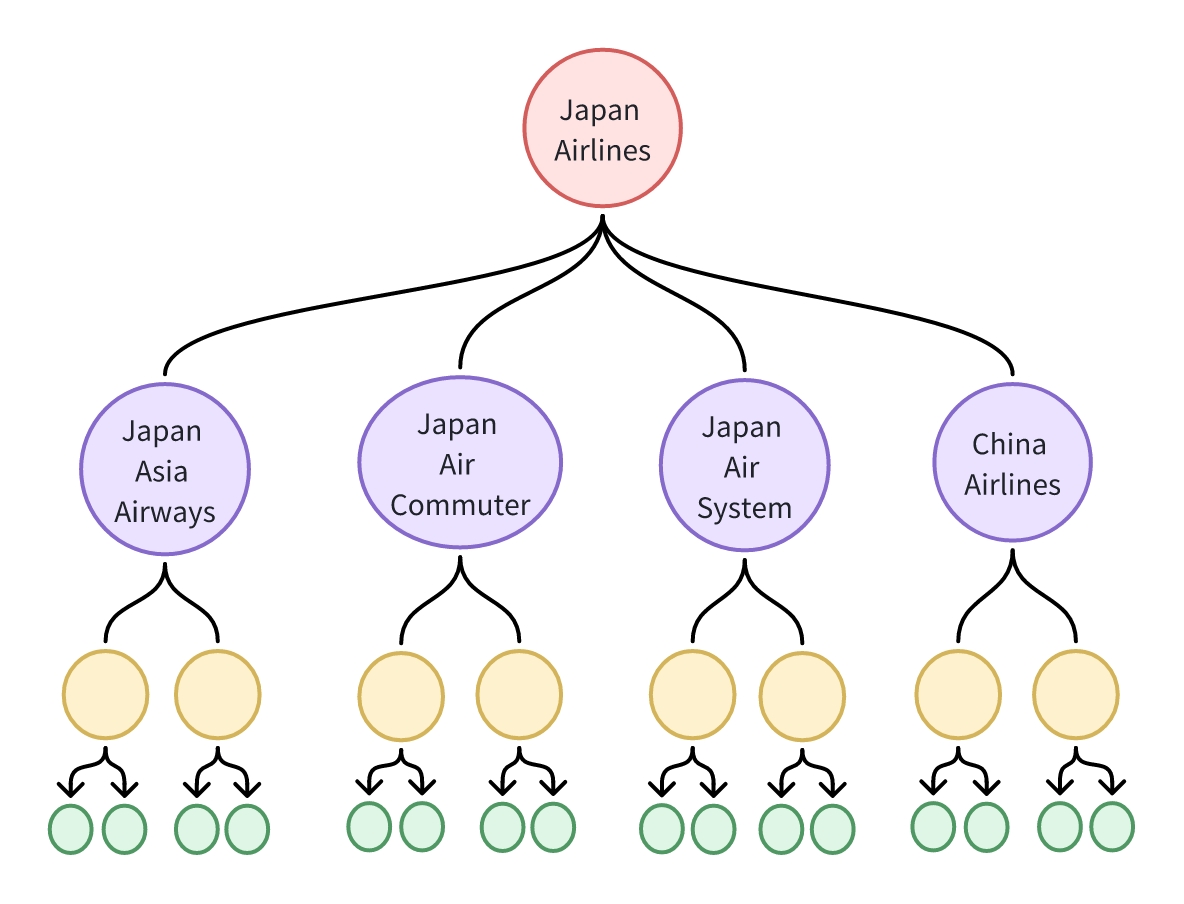}
    \caption{Evidence graph generated by semantic similarity expansion from “Japan Airlines.” The graph shows severe semantic homogeneity, with all expansions remaining within the airline domain and lacking logical diversity.}
  \end{subfigure}
  \hfill
  \begin{subfigure}[t]{0.48\textwidth}
    \centering
    \includegraphics[width=\linewidth]{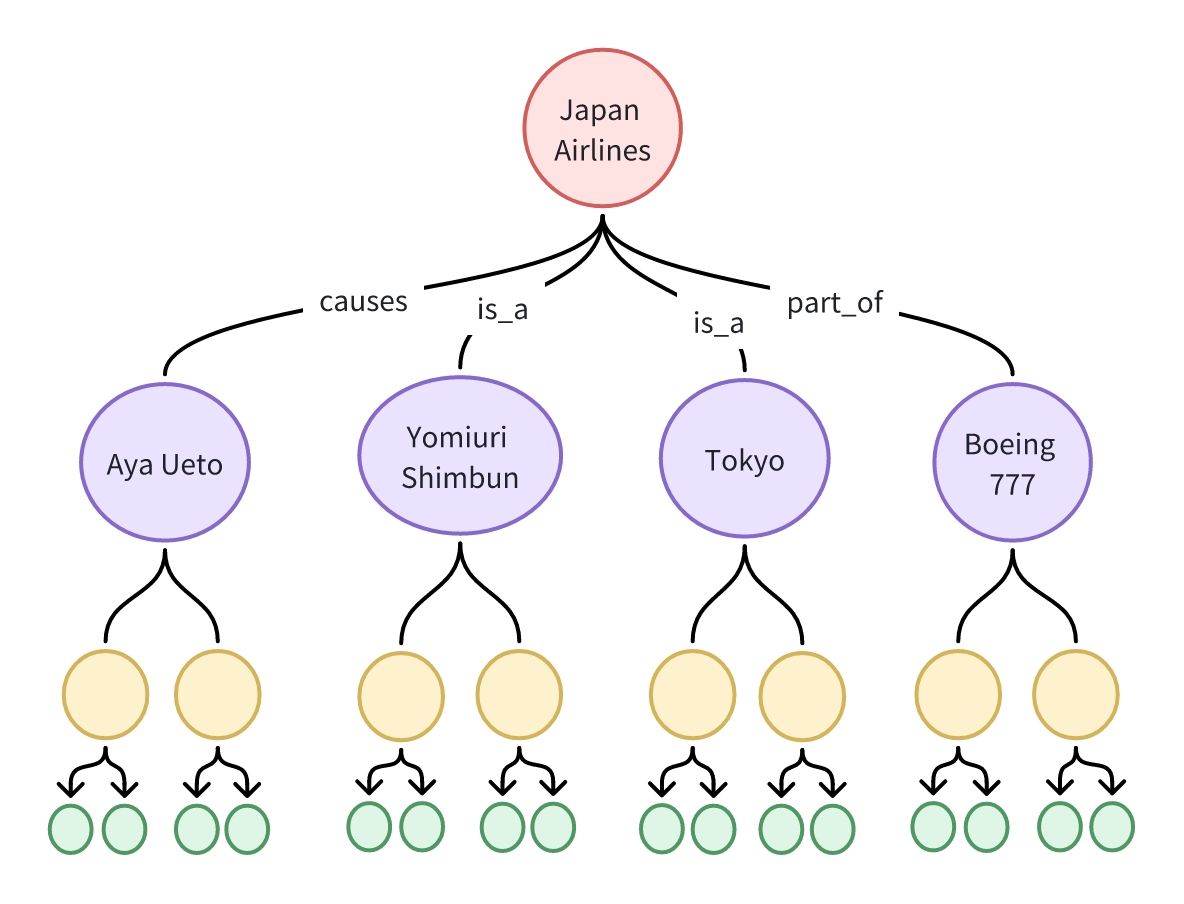}
    \caption{Evidence graph generated from the same seed entity(“Japan Airlines”) after incorporating NLI-based relation classification, logical diversity, and semantic diversity constraints. }
  \end{subfigure}
  \captionsetup{font=small, labelfont=it}
  \caption{A comparative visualization of the semantic-only graph and the relation-augmented graph.}
  \label{fig:relationgraph}
\end{figure}

To overcome this limitation, our approach incorporates \textbf{relation classification} and \textbf{diversity constraints}, ensuring that expansions move beyond surface similarity. With these mechanisms, the evidence cluster evolves into a semantically diverse and logically coherent graph, where entities span multiple categories such as people, locations, organizations, and cultural events. A comparative visualization of the semantic-only graph and the relation-augmented graph is provided in Figure \ref{fig:relationgraph}.

\subsubsection{From Similarity to NLI-Based Relations}
\label{sec: NLI-based_relations}
Early experiments with vector embeddings (e.g., BAAI BGE-M3 \cite{chen2024bge}) encoded entity descriptions and selected expansions by cosine similarity. While effective in grouping thematically related entities, this approach lacked explanatory depth and failed to produce multi-hop reasoning paths.

To overcome this, we adopt a \textbf{Natural Language Inference (NLI)} \cite{sainz2021label} framework. Instead of asking whether two entities are similar, we ask whether an evidence passage entails a hypothesized relation between them. Using the \textit{facebook/bart-large-mnli} \cite{lewis2019bart} model in a zero-shot setting, relation classification is reframed as an entailment task:

\begin{itemize}
    \item \textbf{Premise (P)}: A sentence from webpage containing both entities.
    \item \textbf{Hypothesis (H)}: A templated statement describing a candidate relation.
\end{itemize} 

If the model predicts that P entails H, then the relation is accepted, and the entailment probability serves as its confidence score.
Relations are categorized into six logical types, each with predefined hypothesis templates, as shown in Table~\ref{tab:relation_types} and an example described in Example 2.

    \begin{table}[h]
        \centering
        \renewcommand\cellgape{}
        \caption{Predefined Relation Types Used for NLI-Based Classification}
        \label{tab:relation_types}
        \small
        \begin{tabular}{>{\centering\arraybackslash}m{1.5cm}
                        >{\centering\arraybackslash}m{2.3cm}
                        m{3.5cm}}    
        \toprule
        \textbf{Relation Type} & 
        \textbf{Description} & 
        \multicolumn{1}{c}{\textbf{Example Template}} \\
        \midrule
        \texttt{causes} & Causal relation & \makecell[l]{"{U} causes {V}" \\ "{U} leads to {V}" \\ "{U} induces {V}"} \\ 
        \midrule
        \texttt{part\_of} & Compositional & \makecell[l]{"{U} is part of {V}" \\ "{U} belongs to {V}" \\ "{U} is a component of {V}"} \\
        \midrule
        \texttt{is\_a} & Taxonomic & \makecell[l]{"{U} is a kind of {V}" \\ "{U} is a type of {V}" \\ "{U} is an instance of {V}"} \\
        \midrule
        \texttt{has\_attribute} & Attributive & \makecell[l]{"{U} has attribute {V}" \\ "{U} has property {V}" \\ "{U} is characterized by {V}"} \\ 
        \midrule
        \texttt{requires} & Conditional & \makecell[l]{"{U} requires {V}" \\ "{U} needs {V}" \\ "{U} depends on {V}"}\\ 
        \midrule
        \texttt{used\_for} & Functional & \makecell[l]{"{U} is used for {V}" \\ "{U} is used to access {V}" \\ "{U} serves the purpose of {V}"} \\ 
        \bottomrule
        \end{tabular}
    \end{table}

\begin{tcolorbox}[colback=gray!5,colframe=black,title=Example 2: Japan Airline - Shingo Katori]
Consider the relation between \textit{Japan Airlines} and \textit{Shingo Katori} (a Japanese actor and member of SMAP).
\begin{itemize}
    \item \textbf{Premise}: “At the end of 2005, Japan Airlines began using a Boeing 777 (JA8941) featuring Japanese actor Shingo Katori on one side, and the television series Saiyūki on the other.”
    \item \textbf{Hypotheses}: Generated across all relation templates, in both directions (U→V and V→U).
    \item \textbf{Result}: The NLI model predicted highest confidence (0.92) for “Shingo Katori is an attribute of Japan Airlines.”
    \item \textbf{Decision}: The relation is classified as has\_attribute, direction = backward, confidence = 0.92.
\end{itemize} 

This edge no longer represents vague similarity, but a \textbf{precise semantic link}: Japan Airlines (via aircraft livery) was associated with Shingo Katori as a cultural symbol.

\end{tcolorbox}

\subsubsection{Graph Expansion: Controlled Breadth-First Growth}

The construction of evidence chains proceeds through a \textbf{controlled breadth-first expansion} \cite{cormen2009introduction} strategy. Unlike depth-only approaches that risk producing monotonous, semantically narrow chains, our method ensures that expansions are both logically valid and semantically diverse. The process integrates relation judgment, diversity-aware scoring, and greedy node selection into a unified workflow.  

The part 3 of the Figure~\ref{fig:wf_construction} provides a schematic overview of this pipeline, showing how relation classification guarantees the validity of edges, diversity evaluation prevents redundancy, and breadth-first expansion yields a balanced and interpretable evidence cluster.  


The algorithm operates layer by layer, starting from a seed entity node and expanding outward according to a user-specified strategy (e.g., $[4, 2, 2]$, meaning four nodes at depth 1, two per node at depth 2, and so on). At each step, three main components govern the expansion:

\begin{enumerate}
    \item \textbf{Candidate Sampling}  
    \begin{itemize}
        \item Remove nodes already present in the graph to avoid cycles.  
        \item Perform frequency-based sampling: 70\% high-frequency entities from the source text, combined with 30\% randomly chosen candidates to preserve exploration.  
    \end{itemize}

    \item \textbf{Relation Evaluation}  
    \begin{itemize}
        \item Retrieve evidence sentences containing both parent and candidate entities.  
        \item Apply NLI-based relation classification (see Section~3.2) to assign a relation type, direction, and confidence score.  
        \item Discard edges below the confidence threshold (e.g., 0.45).  
    \end{itemize}

    \item \textbf{Diversity-Aware Scoring}\\
    Each candidate is assigned a composite score that integrates logical and semantic considerations:  
    \begin{equation}
        \begin{split}
        \text{Score} = & \ \alpha \cdot \text{NLI Confidence} \\
        & + \beta \cdot \text{Relation Diversity} \\
        & + \gamma \cdot \text{Semantic Diversity}
        \end{split}
    \end{equation}

    \begin{itemize}
        \item \textit{Relation diversity:} Penalize repeated relation types within the same layer.  
        \item \textit{Semantic diversity:} Penalize nodes whose titles are too similar to already selected nodes.  
        \item \textit{Paragraph diversity:} Favor nodes extracted from different sections of the page, ensuring broader contextual coverage.  
    \end{itemize}

    \item \textbf{Greedy Selection and Expansion}  
    \begin{itemize}
        \item Rank all candidates by their composite scores.  
        \item Select the top-K candidates as the new child nodes.  
        \item Add these nodes and their corresponding edges (including evidence passage, relation type, direction, and confidence) into the graph.  
    \end{itemize}
\end{enumerate}

The final result of this expansion process is a \textbf{graph object} that records the layered evidence cluster in a structured form. Each node stores standardized entity information (\texttt{ID}, \texttt{title}, \texttt{type}), while each edge stores its associated \texttt{evidence passage}, \texttt{relation type}, \texttt{direction}, and \texttt{confidence score}. The entire graph is serialized into a JSON format (\texttt{GraphData}), making it both human-interpretable and machine-readable.  

For example, starting from the seed entity \textit{Japan Airlines}, the expansion yields a multi-layered graph that links not only to other airlines, but also to related people (e.g., Shingo Katori), organizations (e.g., SMAP), locations (e.g., Tokyo, Sapporo), and cultural events (e.g., Kōhaku Uta Gassen). In addition, the system enriches the seed entity with discriminative attributes—such as founding year (1951), hub (Osaka–Kansai), and alliance membership (Oneworld)—so that these attribute cues alone can uniquely identify the correct seed entity (Japan Airlines). This demonstrates that the system avoids semantic homogeneity, instead producing logically coherent and semantically diverse chains that serve as high-quality raw material for downstream multi-hop question generation.

\subsection{Question Construction \& Optimization}
The final stage of our pipeline transforms the evidence clusters generated in Step 3 into \textbf{complex multi-hop questions}. The input consists of the \textbf{evidence cluster} (nodes, edges, and supporting passages) and the \textbf{seed entity}. Unlike naive question generation, which often yields factoid-style queries easily solved by direct retrieval, our method employs a \textbf{bottom-up, reverse reasoning strategy}. This ensures that the constructed question faithfully reproduces the reasoning path encoded in the evidence cluster.

\subsubsection{Reverse Question Generation}
We adopt a bottom-up approach that begins from the \textbf{leaf nodes} of the evidence cluster and progressively backtracks to the seed entity node. At each step, the model is prompted to convert local evidence into a \textbf{descriptive but oblique clue}, which avoids explicitly mentioning the seed entity or its immediate neighbors. These clues are then recursively nested to form a \textbf{composite reasoning chain}, culminating in a single question that uniquely points to the \textbf{seed entity}.

To guarantee difficulty and uniqueness, the generation process enforces several constraints:

\begin{itemize}
    \item Each question must integrate at least $n$ deep cues, where $n$ is a configurable parameter depending on the project setting. Here, $depth$ is defined as the number of edges between the seed entity (the answer node) and the evidence node in the underlying evidence graph. This ensures that questions require reasoning beyond shallow retrieval.
    \item Preference is given to cues drawn from deeper nodes rather than immediate neighbors ($depth=1$), reducing the likelihood of trivial single-hop retrieval.
    \item Descriptions favor \textbf{category–relation–event} expressions over surface names (e.g., “a major East Asian carrier” instead of “Japan Airlines”).
\end{itemize} 

\subsubsection{Question Obfuscation}
Once an initial draft is generated, the question undergoes a controlled obfuscation process. Explicit anchors such as exact years, city names, or institutional titles are generalized into higher-level descriptors (e.g., “in the early 21st century” instead of “2003”; “a North American diplomatic authority” instead of “the U.S. Secretary of State”). Similarly, personal names are replaced by roles or contextualized identifiers (e.g., “an artist who specialized in landscape paintings featuring misty and rainy scenes” instead of the exact name). This abstraction raises the difficulty of direct retrieval while preserving solvability.

\subsubsection{Iterative Refinement}
To further ensure robustness, the system performs automatic question optimization. After optimization, multiple large language models attempt to answer the constructed question. If a majority of models successfully identify the seed entity, the system triggers a \textbf{rewriting procedure} that systematically increases the reasoning difficulty while preserving answer uniqueness.

This rewriting procedure is grounded in the underlying evidence graph. First, the system selects a minimal supporting subgraph containing multiple deep cues (i.e., nodes located several hops away from the seed entity, beyond the most immediate neighbors). To prevent shallow retrieval, anchor terms and “killer pairs” (e.g., {award + year}, {model + maker}) are softened or replaced with implicit formulations. Rather than directly naming entities, the system rephrases them using higher-level categories, relations, and event descriptors, ensuring that no direct level-1 node or alias appears in the surface text.

The rewriting process follows a structured pipeline:

\begin{enumerate}
    \item Graph parsing and subgraph selection (core axis + deep cues);
    \item Text generation using implicit and oblique formulations;
    \item Self-verification, including checks on the number and depth of cues, logical coherence, strict uniqueness, alias avoidance, and length constraints;
    \item If any constraint fails, the system rolls back and regenerates the question.
\end{enumerate}

Through this loop, the discriminative properties required to uniquely identify the seed entity are preserved, but the search complexity is progressively increased. This ensures that even as the question becomes harder, the seed entity remains the only correct answer. The loop terminates once the question reaches the target difficulty level or the maximum number of refinement rounds.

\section{Data Quality Evaluation System}
To ensure that automatically constructed questions are not only of intended difficulty, but also solvable and unique, we design a Data Quality Evaluation System that serves as a filtering layer between question generation and final dataset inclusion. It consists of two complementary components: a \textbf{graph-based textual structure} (Section \ref{sec:graph-based_textual_structure}) that provides early structural screening based on linguistic and reasoning features, and a \textbf{data quality evaluation workflow} (Section \ref{sec:data_quality_evaluation_wf}) that conducts constraint decomposition and evidence-based validation. Together, these components ensure that only structurally coherent, sufficiently complex, and verifiably unique questions enter the final dataset.

\subsection{Graph-Based Textual Structure}
\label{sec:graph-based_textual_structure}
To systematically assess the quality of constructed questions before formal evaluation, we introduce a graph-based textual structure representation. This representation extracts the \textbf{subject}, \textbf{object}, and \textbf{attribute} elements embedded in the constructed question and links them using a predefined set of six linguistic relations (Section \ref{sec: NLI-based_relations}): \textit{is\_a}, \textit{part\_of}, \textit{has\_attribute}, \textit{causes}, \textit{requires}, and \textit{used\_for}. By converting the textual prompt into a structured graph, we make the underlying reasoning chain explicit, auditable, and easily visualizable.

We place this graph-structuring step immediately after question construction but before the formal Data Quality Evaluation (Section \ref{sec:data_quality_evaluation_wf}). This allows us to conduct early-stage structural screening: questions that fail to form coherent and solvable graphs can be automatically discarded before entering the more computationally expensive verification stage. Only structurally valid and potentially high-quality questions flow into the evaluation pipeline.

To assess graph quality, we compute four key structural indicators:
\begin{enumerate}
    \item \textbf{Orphan nodes} — Whether the graph contains nodes that are not connected to the main reasoning chain. 
    \begin{itemize}
        \item Criterion: $\text{OrphanNodes} = 0$ guarantees basic solvability.
    \end{itemize}
    
    \item \textbf{Attribute count ($T_a$)} — The number of attribute nodes attached to the core graph.
    \begin{itemize}
        \item Criterion: $T_a \geq \alpha$, where $\alpha$ is a project-specific threshold. A larger $T_a$ indicates richer semantic specificity, which strengthens answer uniqueness.
    \end{itemize}

    \item \textbf{Edge count ($T_e$)} — The total number of edges in the graph.
    \begin{itemize}
        \item Criterion: $T_e \geq \beta$, where $\beta$ is project-dependent. A higher $T_e$ reflects more complex reasoning paths and multi-constraint composition.
    \end{itemize}

    \item \textbf{Graph diameter ($T_d$)} — The longest shortest path between any two nodes in the graph.
    \begin{itemize}
        \item Criterion: $T_d \geq \gamma$, where $\gamma$ is a tunable threshold. A larger diameter indicates deeper multi-hop reasoning chains.
    \end{itemize}
\end{enumerate}

The thresholds $\alpha$, $\beta$, and $\gamma$ can be flexibly set according to the difficulty level required by specific projects. Absence of orphan nodes ensures basic solvability, while sufficiently large $T_e$ and $T_d$ indicate reasoning complexity, and a higher $T_a$ enforces uniqueness.

\begin{figure*}[t]
    \centering
    \includegraphics[width=\textwidth]{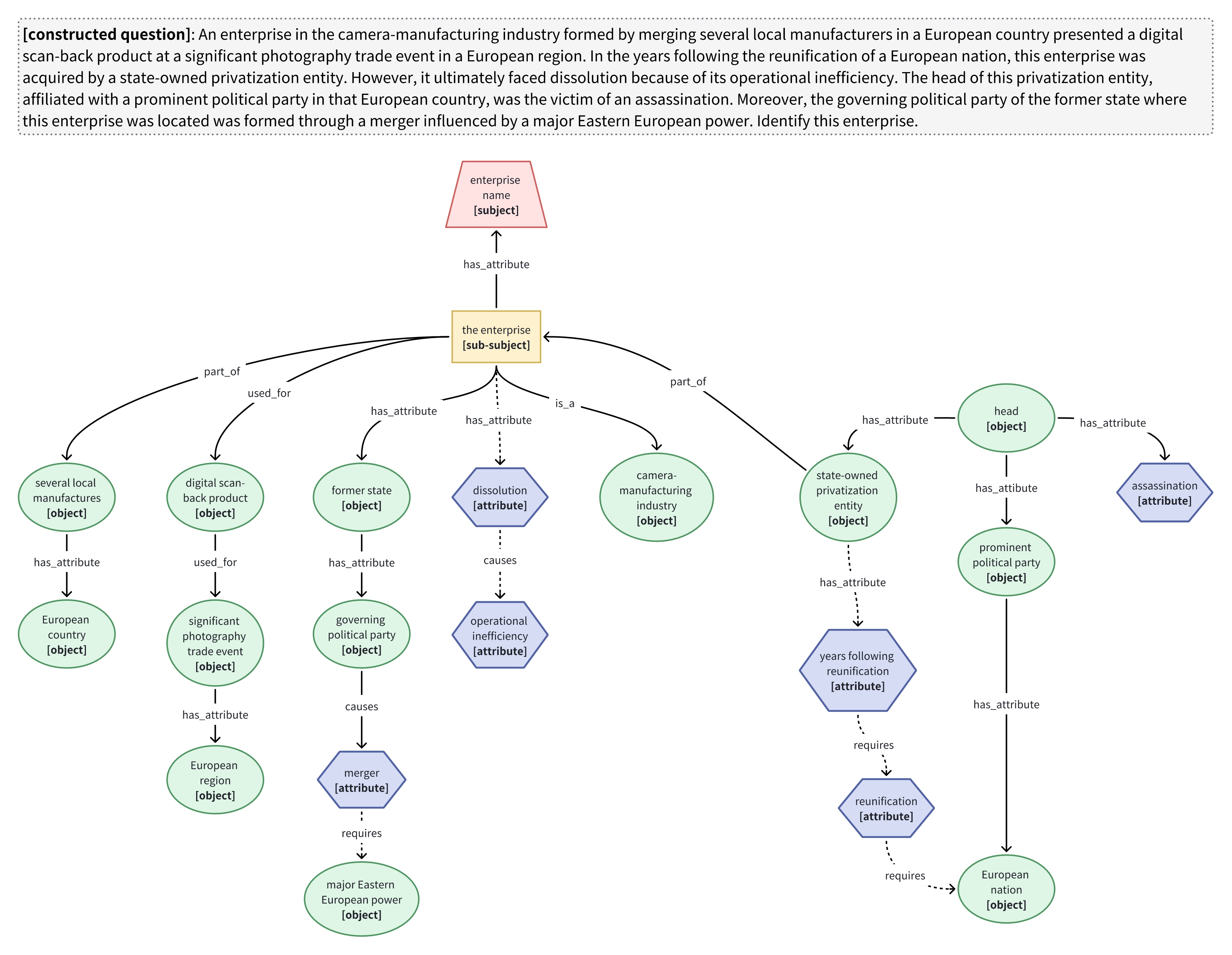}
    \captionsetup{font=small, labelfont=it}
    \caption{Example of graph-based textual structure extracted from a constructed multi-hop question. Nodes represent subjects, objects, and attributes, while edges encode linguistic relations (is\_a, part\_of, has\_attribute, causes, requires, used\_for).}  
    \label{fig:graph_based_textual_structure}
\end{figure*}

An example of a well-structured graph extracted from a constructed question is shown in Figure \ref{fig:graph_based_textual_structure}. This example illustrates how a linguistically complex prompt can be decomposed into a structured graph of entities, attributes, and relations, forming the foundation for subsequent quality verification.

\subsection{Data Quality Evaluation Workflow}
\label{sec:data_quality_evaluation_wf}
To ensure the reliability and discriminative power of the constructed dataset, we design a two-step Data Quality Evaluation Workflow, as shown in Figure \ref{fig:data_quality_evaluation}. This system integrates multi-model answering, structured clue extraction, and evidence-based validation. The first step quickly eliminates invalid questions through majority voting, while the second step provides a rigorous, auditable evaluation based on explicit clue decomposition and structured evidence matching. As a result, only those questions that are both uniquely solvable and well-supported by evidence are preserved in the final dataset.

\begin{figure*}[t]
    \centering
    \includegraphics[width=\textwidth]{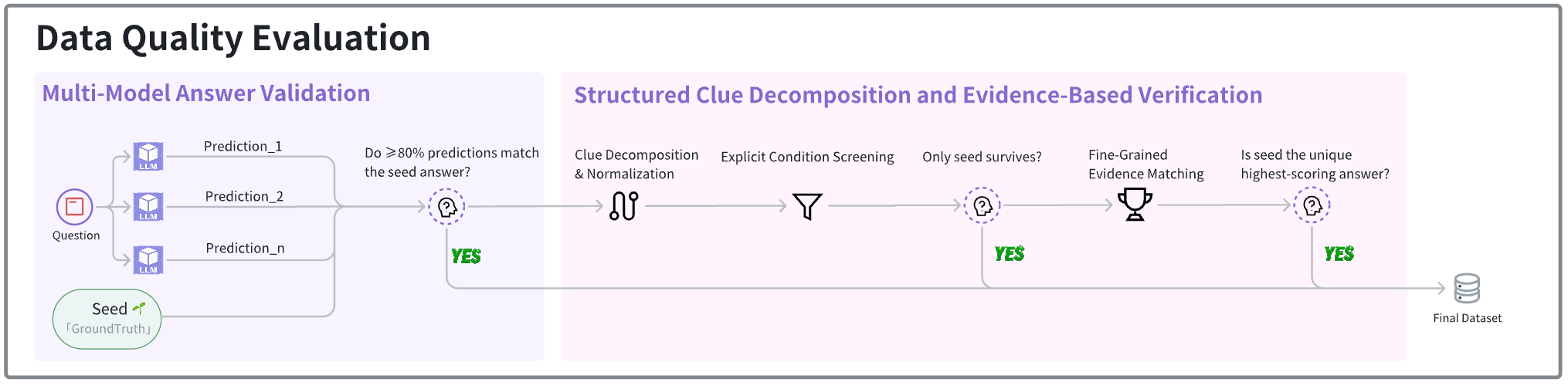}
    \captionsetup{font=small, labelfont=it}
    \caption{A two-step data quality evaluation pipeline. The evaluation process consists of: (1) \textbf{Multi-Model Answer Validation}, and (2) \textbf{Structured Clue Decomposition \& Evidence-Based Verification}.}  
    \label{fig:data_quality_evaluation}
\end{figure*}

\subsubsection{Prediction-Based Validation}
In the first stage, the constructed question is answered by multiple inference runs, yielding a set of candidate predictions $\{\text{seed entity}, \text{prediction1}, \text{prediction2}, \dots\}$. If the majority of predictions match the seed entity, the question is considered stable and accepted directly. Otherwise, it is flagged for further analysis, together with the seed entity and all alternative predictions.
This serves as a coarse filtering mechanism, effectively removing questions with unclear or inconsistent reasoning.

\subsubsection{Constraint Decomposition and Evidence-Based Verification}
For questions that fail the initial validation, we apply a structured verification procedure composed of three tightly coupled steps:
\begin{enumerate}
    \item \textbf{Clue Decomposition and Normalization}

    The question is decomposed into a set of atomic, verifiable constraints expressed as structured predicates.

    \begin{itemize}
        \item Vague time expressions are normalized into explicit intervals (e.g., “early 2020s” $\rightarrow$ $[2020,2023]$).
        \item Fuzzy geographic or categorical terms are mapped into structured hints (e.g., “southern Indian state” $\rightarrow$ $\{\text{region\_hint: South}\}$).
        \item Each predicate is annotated with its source span and a confidence score, while uncertain values are represented as null with an accompanying explanation.
    \end{itemize}
    
    This yields a structured representation that ensures full coverage of the question text.

    \item \textbf{Explicit Condition Screening}

    The structured predicates are used to evaluate each candidate answer against explicit conditions such as time, location, entity type, and key attributes.

    \begin{itemize}
        \item Candidates that fail to satisfy any explicit constraints are discarded immediately.
        \item If only the seed entity remains, the question is accepted at this stage.
        \item If multiple candidates remain, they are ranked based on their explicit-match score $S_{\text{exp}}$. If the seed entity achieves the uniquely highest score, the question is accepted; otherwise, it moves to fine-grained verification.
    \end{itemize}

    \item \textbf{Fine-Grained Evidence Matching}

    For remaining candidates, we perform a detailed comparison against an evidence pack built from local knowledge bases and, if necessary, external retrieval.

    \begin{itemize}
        \item Each predicate is evaluated under a four-value scheme: Y (Match), P (Partial), U (Unknown), N (Contradiction).
        \item High-priority contradictions (N) result in immediate elimination.
        \item A normalized score $S_{\text{norm}}$ is computed by aggregating weighted evaluations across all constraints.
        \item Explicit evidence references and concise justifications are required for each decision.
        \item If the seed entity achieves the uniquely highest $S_{\text{norm}}$, the question is retained; otherwise, it is discarded.
    \end{itemize}
\end{enumerate}

\section{Experiments}
To evaluate both the difficulty and answerability of our automatically constructed multi-hop questions, we conducted a two-part experimental study. The first part measures model performance across several representative datasets to assess whether our generated questions achieve comparable reasoning depth to real-world retrieval-intensive benchmarks. The second part performs a graph-based structural analysis to quantify the reasoning and relational complexity of each dataset through linguistic and topological metrics.

\subsection{Model Evaluation on Multi-Hop Datasets}
We evaluated multiple large language models -- GPT-5 (Advanced Reasoning), Gemini-2.5-Pro, Doubao-1.6-Thinking, and Deepseek-V3.2 -- on four datasets: HotpotQA, ComplexWebQuestions, BrowseComp, and our BMGQ dataset. Each model independently generated answers, and correctness was measured by strict exact-match comparison against reference answers. The overall results are summarized in Table~\ref{tab:model_comparison}:

\begin{table}[htbp]
\centering
\caption{Model performance comparison across datasets.}
\begin{tabular}{lcccc}
\toprule
\textbf{Dataset / Model} & \textbf{GPT-5} & \textbf{Gemini-2.5-Pro} & \textbf{Doubao-1.6-Thinking} & \textbf{Deepseek-V3.2} \\
\midrule
\textbf{HotpotQA} & 76\% & 49\% & 43\% & 39\% \\
\textbf{ComplexWebQuestions}    & 55\% & 56\% & 50\% & 33\% \\
\textbf{BrowseComp}    & \textbf{23\%} & \textbf{12\%} & \textbf{1\%} & \textbf{1\%} \\
\textbf{BMGQ Dataset}     & \textbf{37\%} & \textbf{13\%} & \textbf{5\%} & \textbf{3\%} \\
\bottomrule
\end{tabular}
\label{tab:model_comparison}
\end{table}

Across all models, accuracy drops sharply from HotpotQA and ComplexWebQuestions to BMGQ dataset and BrowseComp, revealing a substantial increase in reasoning and retrieval difficulty. While HotpotQA primarily involves shallow two-hop reasoning, both BMGQ and BrowseComp dataset emphasize hard-to-search but easy-to-verify questions -- where individual clues remain ambiguous, but their combination uniquely constrains the answer space.

GPT-5 maintains the highest overall accuracy, yet its performance on our BMGQ dataset (37\%) remains close to that on BrowseComp (23\%), confirming that our automatically generated questions approach evaluation-level difficulty. Meanwhile, models such as Gemini-2.5-Pro and Doubao-1.6-Thinking show particularly steep performance declines, underscoring that these datasets demand multi-domain reasoning and longer retrieval chains, which exceed the capacity of models primarily tuned for short-hop factual QA.

Overall, the results demonstrate that our automatic construction framework successfully produces questions comparable in complexity to leading evaluation datasets while retaining scalability for training use.

\subsection{Graph-based Structural Analysis}
To characterize the structural and semantic complexity of different multi-hop QA datasets, we construct a graph representation for each question, as described in Section~\ref{sec:graph-based_textual_structure}. Each question is converted into a directed heterogeneous graph (G = (V, E)), where nodes represent objects (entities) or attributes (properties, events, conditions), and edges represent one of six linguistically grounded relation types. We compute a set of graph-theoretic and semantic metrics over these graphs to quantify dataset difficulty along the axes of breadth, depth, constraint richness, and relational complexity. Below, we detail each metric, its computation, and the capability dimension it captures.

\medskip
\noindent\textbf{(1) Objects (\#).}
\begin{itemize}[leftmargin=2em]
    \item[] \textit{Definition.} Number of object nodes extracted from the prompt.
    \item[] \textit{Interpretation.} Measures the breadth and depth of the reasoning space: more objects imply that the question touches more entities, domains, or facts, increasing the search space and making shortcut retrieval less viable.
    \item[] \textit{How it is obtained.} Directly counted from the graph’s object-type nodes.
\end{itemize}

\medskip
\noindent\textbf{(2) Attributes (\#).}
\begin{itemize}[leftmargin=2em]
    \item[] \textit{Definition.} Number of attribute nodes representing properties, states, temporal descriptors, or contextual modifiers.
    \item[] \textit{Interpretation.} Reflects constraint richness: more attributes impose stricter filtering conditions and help enforce answer uniqueness.
    \item[] \textit{How it is obtained.} Count the attribute-type nodes extracted during relation parsing.
\end{itemize}

\medskip
\noindent\textbf{(3) Number of Edges.}
\begin{itemize}[leftmargin=2em]
    \item[] \textit{Definition.} Total number of undirected relations among nodes in (E).
    \item[] \textit{Interpretation.} Higher edge density typically indicates more interdependencies between clues, which increases reasoning entanglement and reduces the risk of shallow single-hop resolution.
\end{itemize}

\medskip
\noindent\textbf{(4) Average Degree.}
\begin{itemize}[leftmargin=2em]
    \item[] \textit{Definition.} $\text{AvgDegree}(G) = \frac{1}{|V|}\sum_{v \in V} \deg(v)$
    \item[] \textit{Interpretation.} Captures the overall interconnectedness of the semantic structure.
 A higher value suggests that nodes participate in multiple relations, increasing combinatorial complexity.
    \item[] \textit{How it is computed.} Degree is counted in an undirected sense to reflect relational connectivity regardless of linguistic directionality.
\end{itemize}

\medskip
\noindent\textbf{(5) Maximum Degree.}
\begin{itemize}[leftmargin=2em]
    \item[] \textit{Definition.} $ \text{MaxDegree}(G) = \max_{v \in V} \deg(v)$
    \item[] \textit{Interpretation.} Indicates the presence of hub nodes, often representing pivotal clues that connect multiple reasoning branches. Datasets with higher max-degree tend to involve more intricate, non-linear reasoning.
\end{itemize}

\medskip
\noindent\textbf{(6) Graph Diameter.}
\begin{itemize}[leftmargin=2em]
    \item[] \textit{Definition.} Longest shortest-path distance between any pair of nodes in the undirected graph.
    \item[] \textit{Interpretation.} Measures the depth of reasoning: larger diameters correspond to longer chains of dependencies and more multi-step inference.
\end{itemize}

\medskip
\noindent\textbf{(7) Longest Path Starting from the Subject.}
\begin{itemize}[leftmargin=2em]
    \item[] \textit{Definition.} Length of the longest simple path originating from the designated subject node.
    \item[] \textit{Interpretation.} Captures targeted depth: how far a solver must propagate from the main subject to integrate peripheral clues.
    \item[] \textit{How it differs from diameter.} Diameter measures global structure; this metric isolates subject-centered reasoning chains.
\end{itemize}

\medskip
\noindent\textbf{(8) Number of Linear Logical Paths.}
\begin{itemize}[leftmargin=2em]
    \item[] \textit{Definition.} Count of simple, non-branching paths in which every intermediate node has degree 2.
    \item[] \textit{Interpretation.} Represents the number of independent reasoning chains embedded in a question. More paths → broader search space and more compositional subproblems.
\end{itemize}

\medskip
\noindent\textbf{(9) Number of Cycles.}
\begin{itemize}[leftmargin=2em]
    \item[] \textit{Definition.} Number of simple cycles in the undirected graph.
    \item[] \textit{Interpretation.} Cycles reflect interdependent reasoning, where clues reinforce each other; datasets with more cycles tend to resist shortcut single-hop strategies.
\end{itemize}

\medskip
\noindent\textbf{(10) Number of Intersection Nodes.}
\begin{itemize}[leftmargin=2em]
    \item[] \textit{Definition.} Nodes with $\text{degree} \ge 3$ serving as junctions connecting multiple reasoning threads.
    \item[] \textit{Interpretation.} A high count indicates branching complexity: solvers must integrate multiple paths rather than follow one linear chain.
\end{itemize}

\medskip
\noindent\textbf{(11) Mean Relation-Type Diversity.}
\begin{itemize}[leftmargin=2em]
    \item[] \textit{Definition.} Average number of distinct relation types (among the six defined in Section~\ref{sec:graph-based_textual_structure}) per question: $\text{RelationDiversity} = \frac{1}{6}\sum_{r} \text{count}(r)$
    \item[] \textit{Interpretation.} Quantifies the balance and distributional evenness of relation types across a question’s evidence graph—datasets with more uniformly distributed causal, taxonomic, compositional, attributive, temporal/locational, and other relations avoid over-reliance on any single relation category and better reflect structurally varied reasoning patterns.
\end{itemize}

\medskip
\noindent\textbf{(12) Undirected Degree Centrality (Score).}
\begin{itemize}[leftmargin=2em]
    \item[] \textit{Definition.} $C_D(v) = \frac{\deg(v)}{|V|-1}$, aggregated across nodes (mean).
    \item[] \textit{Interpretation.} Indicates how influence is distributed across the graph; high centrality suggests concentrated evidence hubs requiring prioritization during reasoning.
\end{itemize}

\medskip
\noindent\textbf{(13) Undirected Betweenness Centrality (Score).}
\begin{itemize}[leftmargin=2em]
    \item[] \textit{Definition.} $C_B(v) = \sum_{s \ne v \ne t} \frac{\sigma_{st}(v)}{\sigma_{st}}$, where ($\sigma_{st}$) is the number of shortest paths between (s) and (t).
    \item[] \textit{Interpretation.} Measures whether reasoning passes through narrow “choke points” — high scores imply pivotal clues that cannot be bypassed.
\end{itemize}

\medskip
\noindent\textbf{(14) Undirected Eigenvector Centrality (Score).}
\begin{itemize}[leftmargin=2em]
    \item[] \textit{Definition.} Centrality proportional to the centrality of neighbors: $x_i = \frac{1}{\lambda} \sum_j A_{ij}\, x_j$.
    \item[] \textit{Interpretation.} Captures global importance: clues connected to other influential clues receive high values. Datasets with elevated eigenvector centrality often encode deeply nested or hierarchical reasoning.
\end{itemize}

To evaluate whether our proposed BMGQ dataset achieves the structural and inference complexity characteristic of challenging multi-hop benchmarks, we further compare its graphical representation with three representative datasets: ComplexWebQuestions, HotpotQA, and BrowseComp. Table~\ref{tab:graph_stats} lists the averages of fourteen structural and semantic metrics. The results show a clear progression in complexity from earlier multi-hop benchmarks to BrowseComp, and then to our BMGQ dataset.

\begin{table}[t]
\centering
\small
\setlength{\tabcolsep}{3pt}
\renewcommand{\arraystretch}{1.1}

\begin{tabular}{lcccc}
\toprule
\textbf{Metric} &
\textbf{ComplexWebQuestions} &
\textbf{HotpotQA} &
\textbf{BrowseComp} &
\textbf{BMGQ} \\
\midrule
Objects & 1.25 & 1.69 & 6.69 & 10.20 \\
Attributes & 0.77 & 1.80 & 6.09 & 6.07 \\
Number of Edges & 2.94 & 4.56 & 14.69 & 17.83 \\
Average Degree & 1.47 & 1.64 & 1.96 & 1.95 \\
Maximum Degree & 2.58 & 3.28 & 6.03 & 6.35 \\
Graph Diameter & 2.29 & 2.98 & 5.42 & 6.84 \\
Longest Path Starting from the Subject & 1.91 & 2.58 & 5.37 & 5.62 \\
Number of Linear Logical Paths & 1.58 & 2.20 & 5.16 & 6.92 \\
Number of Cycles & 0.03 & 0.11 & 0.88 & 0.67 \\
Number of Intersection Nodes & 0 & 0.03 & 0.42 & 0.31 \\
Mean Relation-Type Diversity & 0.49 & 0.76 & 2.47 & 2.95 \\
Undirected Degree Centrality (Score) & 0.92 & 0.77 & 0.46 & 0.39 \\
Betweenness Centrality (Score) & 0.93 & 0.82 & 0.73 & 0.76 \\
Eigenvector Centrality (Score) & 0.68 & 0.64 & 0.59 & 0.59 \\
\bottomrule
\end{tabular}

\caption{Graph-structural metrics.}
\label{tab:graph_stats}
\end{table}

\paragraph{Breadth of reasoning.} 
BMGQ dataset exhibits the largest number of \textit{object} nodes (10.20) among all datasets, nearly doubling that of BrowseComp (6.69) and far exceeding HotpotQA and ComplexWebQuestions ($< 2$). This indicates that BMGQ questions integrate a substantially wider set of entities, covering more domains and forcing the solver to explore a broader search space. The \textit{edge count} follows a similar trend: BMGQ has 17.83 edges on average—significantly larger than BrowseComp (14.69) and an order of magnitude higher than early benchmarks ($< 5$). This demonstrates that BMGQ questions encode more relational interactions and richer factual interdependencies.

\paragraph{Depth and multi-hop reasoning complexity.} 
Both \textit{graph diameter} and \textit{longest path starting from the subject} quantify the depth of reasoning chains. BMGQ shows the highest values (diameter = 6.35; longest subject path = 5.62), closely mirroring BrowseComp (6.03 and 5.37, respectively) and far surpassing HotpotQA or ComplexWebQuestions ($\approx 3$). This reflects that BMGQ, like BrowseComp, requires solvers to traverse long, multi-stage reasoning chains rather than relying on shallow or local evidence.

\paragraph{Branching and compositional complexity.}
The \textit{average degree} (1.95) and \textit{maximum degree} (6.84) of BMGQ again align closely with BrowseComp (1.96 and 6.03), indicating comparably dense local connectivity. Meanwhile, the numbers of \textit{linear logical paths} and \textit{intersection nodes} capture branching structure: BMGQ shows 6.92 linear chains and 0.31 intersection nodes, whereas BrowseComp shows 5.16 and 0.42, respectively. These values together indicate that BMGQ questions encompass multiple parallel reasoning threads with moderate merging complexity -- comparable to BrowseComp yet more structured than early datasets, which exhibit minimal branching ($\approx 0-0.1$ intersection nodes).

\paragraph{Cyclic and interdependent reasoning.}
Cycles reinforce mutual constraints and reduce shortcut-solving. BrowseComp contains 0.88 cycles per question, while BMGQ has 0.67, substantially higher than HotpotQA or ComplexWebQuestions ($\le 0.11$). This shows that BMGQ maintains interlocking logical structures rather than simple linear chains, a hallmark of difficult retrieval-based reasoning tasks.

\paragraph{Semantic richness and relation diversity.}
An important feature of real-world multi-hop reasoning is the diversity of relational semantics. BrowseComp shows an average of 2.47 relation types per question, while BMGQ reaches an even higher diversity of 2.95. In contrast, early datasets employ almost exclusively taxonomic or compositional relations ($< 0.8$). This demonstrates that BMGQ questions integrate a wide mix of attributive, causal, compositional, temporal, and classificatory relations, creating heterogeneous reasoning requirements similar to or exceeding those in BrowseComp.

\paragraph{Centrality-based difficulty.}
The centrality metrics—degree, betweenness, and eigenvector centrality -- characterize how crucial particular nodes are in mediating the flow of reasoning.
\begin{itemize}
    \item In all three centrality scores, BMGQ (0.76/0.76/0.59) nearly matches BrowseComp (0.73/0.46/0.59).
    \item BMGQ’s slightly higher degree and betweenness scores indicate more influential hubs and more mandatory intermediate reasoning steps than in BrowseComp.
    \item Early benchmarks again show inflated degree centrality (due to small graph size) but low betweenness and eigenvector centrality, reflecting shallow and loosely connected structures.
\end{itemize} 

\paragraph{Overall assessment.}
Across all fourteen metrics, BMGQ consistently clusters alongside BrowseComp and diverges sharply from early multi-hop datasets. BMGQ matches or exceeds BrowseComp in:
\begin{itemize}
    \item reasoning breadth (objects, edges),
    \item depth (diameter, longest path),
    \item relation diversity, and
    \item centrality-driven structural difficulty.
\end{itemize} 

Meanwhile, BrowseComp retains slightly higher cyclic and intersection-node complexity, which is expected given its focus on evaluation-only problem design. Nonetheless, the overall structural fingerprints of the two datasets are strongly aligned.

\subsection{Human Quality Assessment}
While model performance and graph-structural analyses jointly demonstrate that our generated questions resemble high-difficulty benchmarks in complexity and reasoning depth, it remains essential to verify the intrinsic quality of the constructed data itself. To this end, we conduct a human evaluation to assess whether the generated questions are well-formed, logically sound, and paired with correct and uniquely identifiable answers.

We randomly sampled a substantial subset of questions from the full dataset and asked human annotators to evaluate them along three dimensions:
\begin{enumerate}
    \item \textbf{Question soundness}: whether the question is well-posed, free of contradictions, and contains no malformed or misleading clues.
    \item \textbf{Answer correctness}: whether the provided seed answer is factually correct and supported by evidence.
    \item \textbf{Answer uniqueness}: whether the clues in the question uniquely identify the seed answer, without allowing alternative plausible candidates.
\end{enumerate}

Annotators evaluated each example by examining the natural-language question together with all explicit constraints stated in the prompt. Their task was to determine whether these constraints, taken collectively, uniquely support the labeled answer (i.e., the seed entity). An item was marked as a failure only if at least one of the following conditions occurred: the question contained contradictory or incoherent cues; the seed entity could not be verified as satisfying all stated conditions; or multiple candidate entities were found to satisfy every constraint in the prompt, indicating lack of answer uniqueness.

Across the sampled subset, the dataset achieved an overall pass rate of 87\%, indicating that the vast majority of automatically constructed questions are logically coherent, correctly labeled, and uniquely solvable. This level of quality is particularly notable given the difficulty of the target task and the minimal amount of human intervention involved. The remaining 13\% of cases typically involved edge scenarios—such as clue over-abstraction or insufficiently restrictive combinations—which can be further alleviated by tightening the constraints in the question optimization stage or adjusting the thresholds in the quality evaluation pipeline.

Together with the strong model-based difficulty evaluation and the alignment of graph-structural metrics with BrowseComp-level complexity, the human assessment confirms that our generated dataset is not only challenging but also reliable, verifiable, and high-quality, validating the effectiveness of the BMGQ framework.

\section{Conclusion}
In this work, we present a comprehensive framework for the automatic construction of multi-hop question answering datasets that bridge the gap between existing benchmarks and real-world reasoning demands. Motivated by the limitations of current datasets—such as the shallow reasoning depth of early benchmarks and the evaluation-only nature of BrowseComp—we aim to generate large-scale, training-ready data that retain the hard-to-search yet easy-to-verify characteristics crucial for deep reasoning evaluation.

To this end, we introduced BMGQ, a Bottom-up Method for Generating Complex Multi-hop Reasoning Questions from semi-structured data. BMGQ comprises a four-stage pipeline encompassing data adaptation, node construction, evidence chain generation, and question synthesis, transforming semi-structured knowledge sources into logically rich, semantically diverse, and structurally coherent reasoning graphs. Furthermore, we propose a Data Quality Evaluation System that integrates multi-model consensus, structured clue decomposition, and evidence-based verification. This system ensures that only questions with unique answers, consistent logic, and verifiable evidence are preserved, significantly improving the reliability and discriminative power of the generated dataset.

By automating the construction of BrowseComp-level multi-hop datasets, BMGQ reduces the prohibitive cost of manual curation while enabling scalable production of challenging, high-quality training data. This advancement has the potential to accelerate research on reasoning-centric large language models and facilitate the development of models capable of complex, multi-step inference.

In future work, we plan to extend this pipeline beyond text-based sources to incorporate multimodal evidence, explore cross-lingual dataset construction, and integrate our framework with reinforcement learning workflows to further enhance model reasoning capabilities. We believe this direction will help bridge the gap between benchmark-oriented evaluation and real-world reasoning applications.

\bibliographystyle{unsrt}   
\bibliography{refs}         

\section*{Appendix}
\subsection*{A. Example of Raw Wikipedia Page Noise}
To illustrate the challenges inherent in processing raw Wikipedia data, we provide an annotated excerpt from the Japan Airlines page (Figure~\ref{fig:wiki_page_noise}). The figure highlights typical sources of semantic noise that must be addressed during preprocessing and candidate filtering:

\begin{itemize}
    \item \textbf{Excessive citations and references}: Sections such as \textit{See also}, \textit{References}, and \textit{External links} introduce numerous links that do not directly contribute to entity relationships.
    \item \textbf{Irrelevant common nouns}: Terms like \textit{tons} or \textit{mail} are common words that, if expanded as graph nodes, would produce meaningless connections.
    \item \textbf{Abstract concepts and verbs}: Words such as \textit{privatised} or \textit{capital} are overly generalized or contextually ambiguous.
    \item \textbf{Weakly related references}: Links embedded in bibliographic or supplementary sections are often tangential and unlikely to contribute to coherent reasoning chains.
\end{itemize}

By systematically removing such elements through preprocessing and Named Entity Recognition (NER)-based filtering, we ensure that only concrete, contextually grounded entities (e.g., All Nippon Airways, Tokyo) remain as candidate nodes for multi-hop reasoning.

\begin{figure}[h]
    \centering
    \renewcommand{\thefigure}{A\arabic{figure}}
    \setcounter{figure}{0}
    \includegraphics[width=\columnwidth]{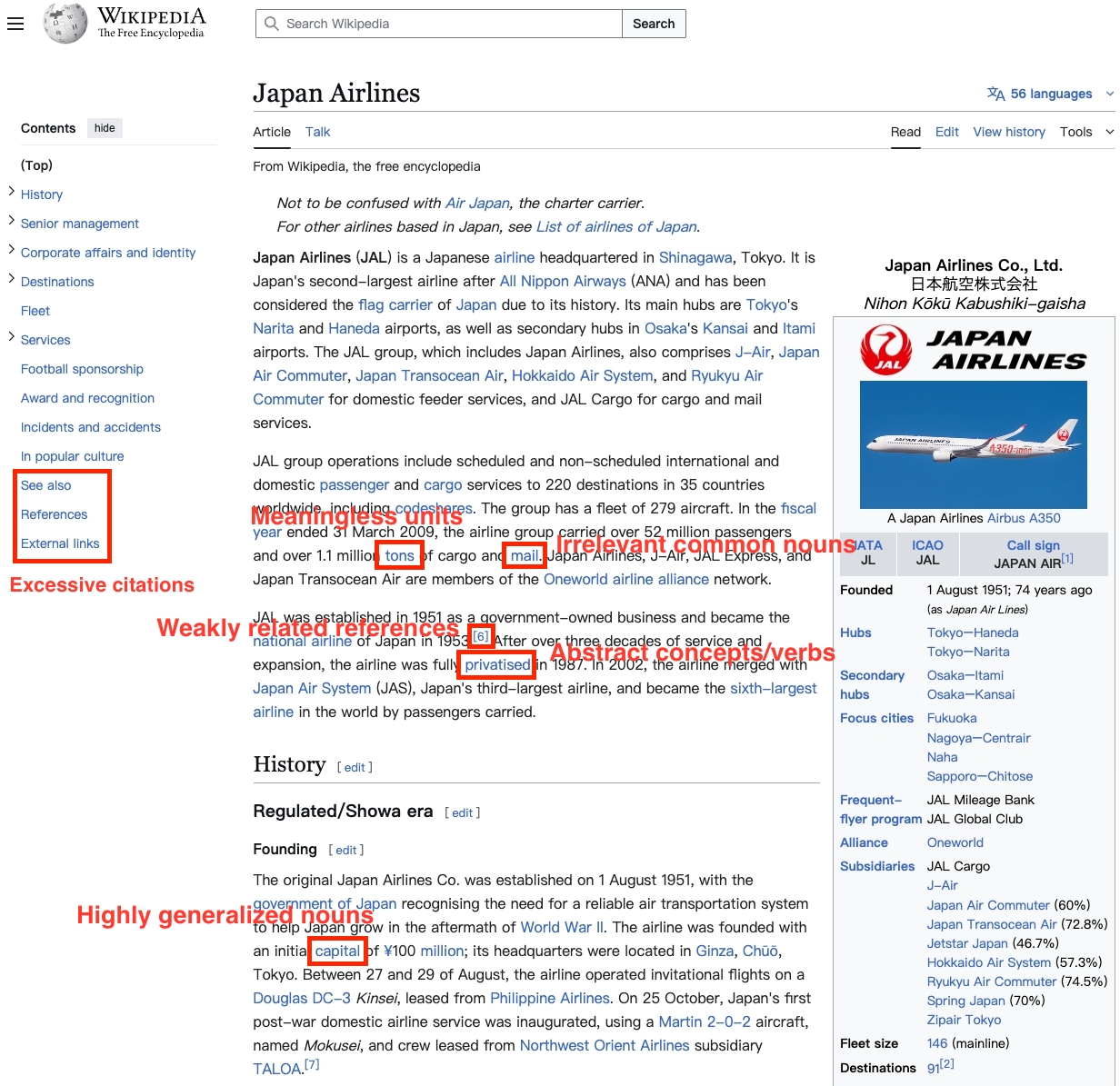}
    \caption{Annotated excerpt from the Wikipedia page “Japan Airlines” highlighting excessive citations, irrelevant common nouns, abstract concepts, and weak references. These noisy elements motivate the preprocessing and filtering steps described in Section 3.2.}
    \label{fig:wiki_page_noise}
\end{figure}

\end{document}